\documentclass[9pt,conference]{IEEEtran}
\IEEEoverridecommandlockouts
\usepackage{cite}
\usepackage{amsmath,amssymb,amsfonts}
\usepackage{algorithmicx}
\usepackage{algorithm}
\usepackage{algpseudocode}
\usepackage{graphicx}
\usepackage{textcomp}
\usepackage{xcolor}
\usepackage{amsmath}
\usepackage{footnote}
\makesavenoteenv{table}
\usepackage{float}
\usepackage{flafter}
\usepackage{array}
\usepackage{multirow}
\usepackage{caption}
\usepackage{subcaption}
\usepackage{fancyhdr}
\usepackage{url}
\usepackage{listings}
\usepackage{fancyvrb}
\usepackage[hidelinks]{hyperref}

\def\BibTeX{{\rm B\kern-.05em{\sc i\kern-.025em b}\kern-.08em
    T\kern-.1667em\lower.7ex\hbox{E}\kern-.125emX}}

\usepackage{graphicx}
\usepackage{eso-pic} 
\newcommand{\PlaceAEBadges}{%
  \AddToShipoutPictureFG*{%
        \put(\LenToUnit{\dimexpr\paperwidth-9cm},\LenToUnit{\dimexpr\paperheight-1.8cm}){%
        {\includegraphics[height=2.1cm]{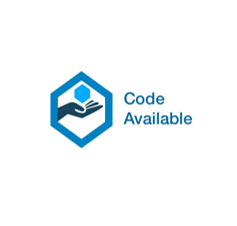}}%
        {\includegraphics[height=2.1cm]{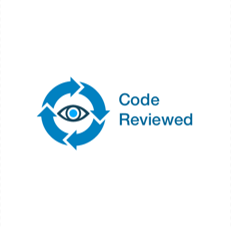}}%
    }%
  }%
}

\begin{document}
\PlaceAEBadges

\title{VCDiag: Classifying Erroneous Waveforms for Failure Triage Acceleration \\
}

\author{\IEEEauthorblockN{Minh Luu}
\IEEEauthorblockA{\textit{Infineon Technologies} \\
Hanoi, Vietnam \\
danhminh.luu@infineon.com}
\and
\IEEEauthorblockN{Surya Jasper, Khoi Le, Evan Pan, Michael Quinn, Aakash Tyagi, Jiang Hu}
\IEEEauthorblockA{\textit{Texas A\&M University} \\
College Station, Texas, USA \\
\{suryajasper,khoile0315,pan.evan,m.d.quinn,tyagi,jianghu\}@tamu.edu}
}


\maketitle


\begin{abstract}
 Failure triage in design functional verification is critical but time-intensive, relying on manual specification reviews, log inspections, and waveform analyses. While machine learning (ML) has improved areas like stimulus generation and coverage closure, its application to RTL-level simulation failure triage, particularly for large designs, remains limited. VCDiag offers an efficient, adaptable approach using VCD data to classify failing waveforms and pinpoint likely failure locations. In the largest experiment, VCDiag achieves over 94\% accuracy in identifying the top three most likely modules. The framework introduces a novel signal selection and statistical compression approach, achieving over 120x reduction in raw data size while preserving features essential for classification. It can also be integrated into diverse Verilog/SystemVerilog designs and testbenches.
\end{abstract}


\begin{figure*}[t]
	\centering
	\includegraphics[width=\textwidth]{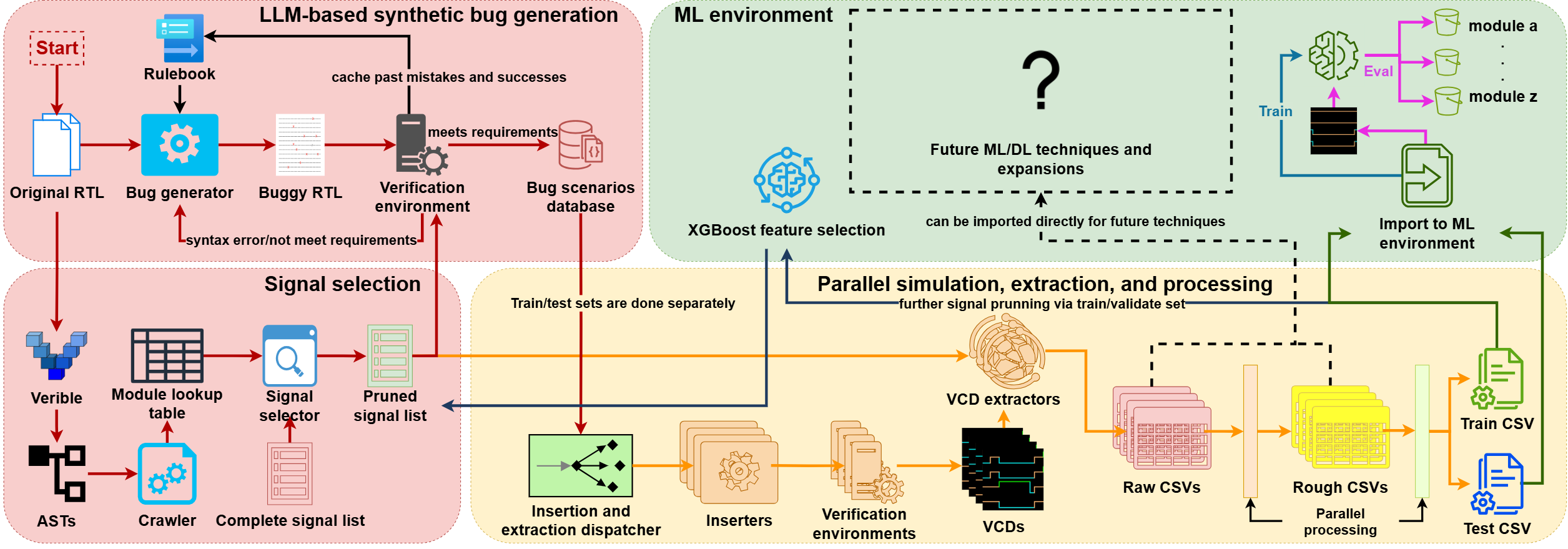}
	\caption{VCDiag framework}
	\label{fig:vcdiag_framework}
\end{figure*}

\section{Introduction}

 Design functional verification (DFV) is critical for ensuring chips meet specifications, avoiding costly errors, and protecting companies' reputations. However, traditional methods struggle with modern design complexities. Verification now consumes more engineering resources than design itself, reflected by a rising verification-to-design engineer ratio~\cite{verifgap2024}. Nearly 75\% of projects miss deadlines and only 14\% avoid a respin due to functional failures~\cite{verifgap2024}. Furthermore, verification engineers spend 47\% of their time debugging~\cite{verifgap2024}. As design complexity grows, innovative debugging tools and methodologies are urgently needed to drive efficiency improvements in verification.

 Failure triage, an important initial step during the debugging process, involves evaluating, prioritizing, and sorting bugs to mitigate critical issues first. At RTL level, coarse-grain debugging is used during the evaluation phase to approximate the root cause’s location, narrowing it down to a likely design module or group of modules. Though pioneering SAT-based techniques~\cite{1512377, 6604054, fae} have demonstrated promising results, their focus on formal verification limits scalability for large IPs and SoCs, where simulation-based verification dominates. ML-powered approaches for simulation rely on log files~\cite{truong2018clustering} or proprietary tools~\cite{safarpour2012failure}, making them dependent on specific testbench configurations and hampering reproducibility. Most importantly, few studies evaluate triage techniques on large RTL designs~\cite{mammo2016bugmd}, and the scarcity of publicly available benchmarks further limits progress toward practical, industry-ready solutions.

 Given these challenges, our work proposes a scalable and testbench-agnostic approach to accelerate coarse-grain debugging and, by extension, failure triage. Our method leverages Value Change Dump (VCD) waveform data, which captures rich temporal and structural design behavior during simulation and is widely supported by industry-standard tools. By mining this data directly, we avoid reliance on proprietary infrastructure or non-robust log parsing. To this end, we present VCDiag, a VCD mining framework for ML-driven classification of failing waveforms to likely RTL modules.

 VCDiag advances existing research by offering three key contributions:

\begin{enumerate}
    
    \item \textbf{A novel approach to failure classification}:  We train ML models on failing waveforms labeled with buggy RTL modules. Once errors are identified in logs, these models help engineers analyze the temporal and structural context within waveforms to infer the most probable failure source, enabling quicker and more accurate coarse-grain debugging.

    \item \textbf{An adaptable and modular infrastructure}: VCD, part of the IEEE Verilog standard~\cite{verilogstandard}, is widely supported by commercial and open-source tools, providing a reliable bridge between verification environments and ML infrastructure. This makes VCDiag adaptable to diverse designs, including those using industry-standard Universal Verification Methodology (UVM) testbenches. The tool is built entirely on open-source libraries~\cite{verible2024, vcdvcd2024, pandas} and evaluated on publicly available designs~\cite{opentitan}~\cite{fabscalar} to support reproducibility and future benchmarking.
  
    \item \textbf{Efficient VCD extraction and processing method}: Large VCD file sizes can make datasets unwieldy, so we propose a multi-step pipeline for efficient VCD extraction and processing. The process begins with an AST-based crawler that builds a lookup table to guide a signal selector in filtering relevant signals for extraction. The extracted data is then trimmed, compressed, and formatted into compact training and testing sets. For large designs, we utilize XGBoost~\cite{xgboost} to rank signal importance and reduce dimensionality. Additionally, the framework supports multi-core parallelization to enhance throughput and efficiency.

\end{enumerate}

 We evaluated VCDiag on advanced RTL designs, including OpenTitan~\cite{opentitan} IPs and the FabScalar processor~\cite{fabscalar}. On the largest benchmark, VCDiag achieves 77\% top-1 and 94\% top-3 accuracy in identifying failure-causing modules. The processing pipeline reduces data size by 123x, from 308GB to 2.5GB, and multi-core parallelism allows simulation, extraction, and processing of 1,600 bug scenarios (over 15,000 VCDs) in under 36 hours on a 32-core system.

\section{Previous Work}

 BugMD~\cite{mammo2016bugmd} detects post-silicon CPU bugs in FabScalar~\cite{fabscalar} by aligning architectural states with a golden reference and extracting mismatch features for bug signature generation. Using synthetic gate-level bug injection and a random forest classifier, it achieves 70\% top-1 and 90\% top-3 accuracy across twelve CPU regions. Since post-silicon timing data and gate-level bug injection are unavailable pre-synthesis, we target a similar problem at the RTL level. We adopted FabScalar’s RTL design, applied comparable metrics, and developed a targeted RTL bug generation tool. Our approach exceeds BugMD’s reported results on FabScalar and consistently performs well across a broader set of modern, open-source designs, demonstrating its scalability and applicability.

\section{Problem Formulation}

 A Verilog/SystemVerilog module defines the behavior of a hardware block through internal logic and signal interactions. Modules communicate via well-defined interfaces and can be composed hierarchically to build complex digital systems. During simulation, the complete design, referred to as the Design-Under-Test (DUT), is exercised by testbenches that apply inputs and check outputs against a golden reference. When a mismatch is detected, the testbench flags the failing cycle and logs diagnostic details. DFV engineers then isolate the failing test case, review the logs, and inspect relevant waveforms against the design specification to identify anomalies and locate the root cause.

\begin{figure}[!h]
	\centering
	\includegraphics[width=0.9\linewidth]{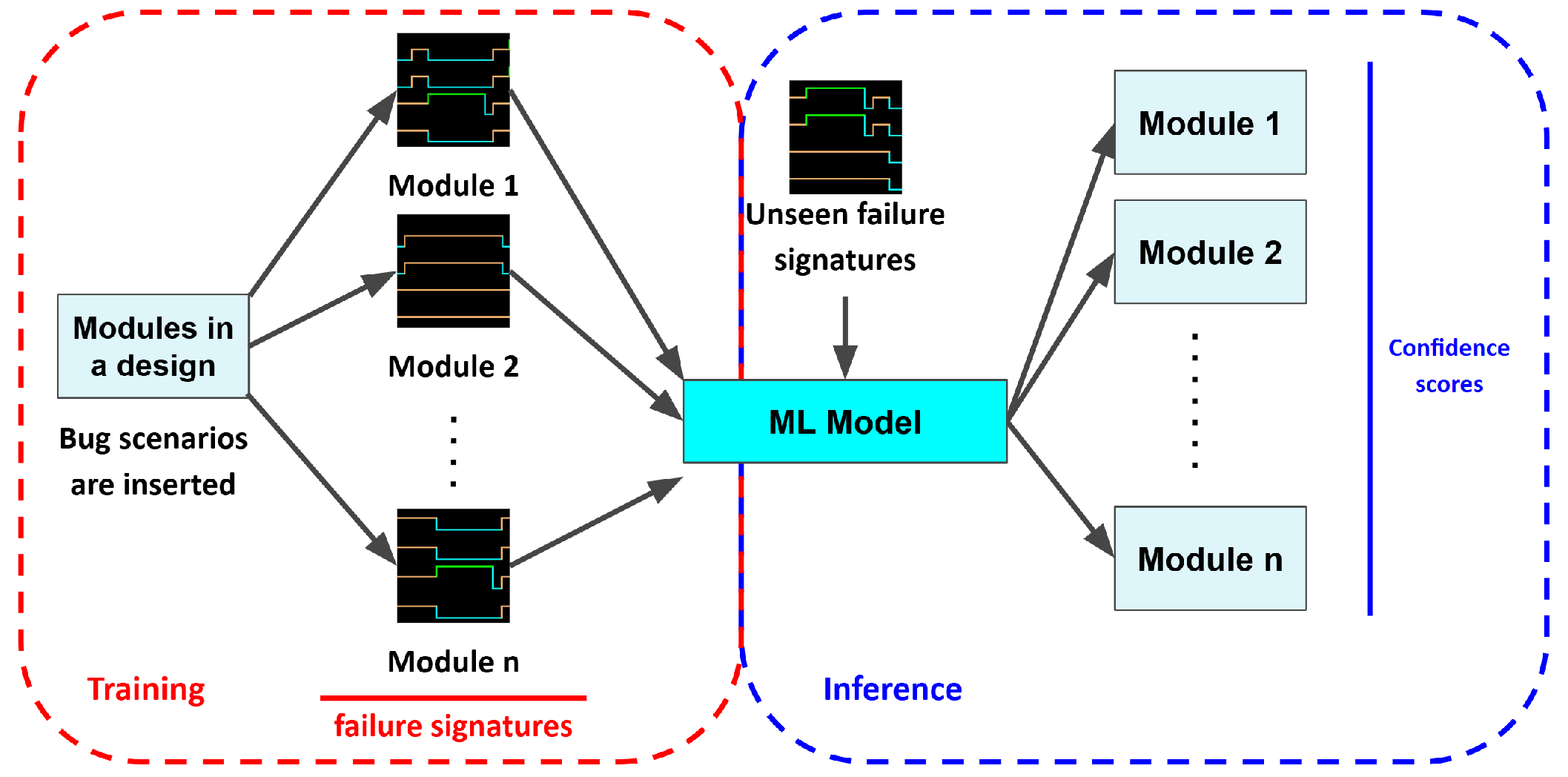}
	\caption{Failure signature classification.}
    \label{fig:big_sig_class}
\end{figure}

 For our analysis, the simulation halts upon detecting the first error, and the resulting VCD (waveform) captures DUT behavior at every simulation tick from initialization to failure. Each VCD is labeled with the module responsible for the failure and serves as a behavioral signature for training ML models as shown in Figure~\ref{fig:big_sig_class}. A separate model is trained and tested for each design to learn its unique behavior. We formulate this as a multivariate time-series classification (MTSC) problem, where a waveform VCD is represented as $X \in \mathbb{R}^{T \times d}$, with $T$ time steps and $d$ signal measurements per step. The objective is to learn a model $f: \mathbb{R}^{T \times d} \to \{1, 2, \dots, M\}$ that maps a failing waveform to the corresponding module label $Y$.

 By using VCDs as the sole input, VCDiag remains independent of the DUT and testbench, ensuring compatibility with any Verilog/SystemVerilog design or simulator without modification. To maintain this generality, future integration of logs or design specs will require standardized outputs or novel feature extraction methods for ML training.

\section{Proposed Framework}

 This section describes the main components of VCDiag in practical execution order, as shown in Figure~\ref{fig:vcdiag_framework}, and concludes with a discussion of the evaluated machine learning models and the rationale for their selection in failing waveform classification.
 
\subsection{Signal selection}

 The primary challenge with VCD is its large size and sparsity, as it stores uncompressed signal activities from the entire design during simulation. This results in millions of data points, making storage and processing highly resource-intensive. Many of these signals are irrelevant to the target modules, and critical events often occur only a few hundred cycles before failure. When triaging failures, engineers typically focus on signals tied to suspected bug sources. In a similar way, we reduce the dataset to scope-relevant signals before training ML models. To do this, we scan each module to capture all internal and interface signals associated with target modules, build a signal lookup table, extract the complete signal list, and prune it to retain only the relevant signals. This targeted selection significantly reduces dataset size, speeds up model training, and improves classification accuracy.

 Analyzing raw RTL code requires extensive parsing effort due to inconsistent formatting styles. To simplify this, we convert RTL into Abstract Syntax Trees (ASTs), which represent the syntactic structure of Verilog/SystemVerilog code. Each node corresponds to a construct such as \texttt{while}, \texttt{if/else}, or \texttt{assign}, and the tree captures hierarchy, variable declarations, control flow, and structure in a standardized format. As long as the code is syntactically valid, the tree structure remains consistent. For each module, we retain submodule instances (\texttt{RegisterVar}'s in the AST) that match our target module list and collect signals from all instances to ensure full coverage.

\begin{algorithm}[h]
\footnotesize
\caption{Build Module Lookup Table from ASTs.}
\label{algo:crawlandbuild}
\begin{algorithmic}[1]
\State \textbf{Input:} ASTs $\alpha$ of design $\delta$
\State \textbf{Output:} Lookup table $\tau$ associating modules $\mu$ with variable types $\sigma$ and variables $\upsilon$
\State Initialize empty table $\tau$
\For{each AST $\alpha$ in $\delta$}
  \For{each $node$ in $\alpha$}
    \If{$node$ is \texttt{ModuleDecl}}
      \State $\mu \gets n.\texttt{NAME}$, $\tau[\mu] \gets \{\}$
      \For{each $s$ in $node.\texttt{SubTree}$}
        \If{$s$ is \texttt{DataType}} 
          \State $\sigma \gets s.\texttt{Type}$, $\tau[\mu][\sigma] \gets \{\}$
        \EndIf
        \If{$s$ is \texttt{RegisterVar}} 
          \State $\upsilon \gets s.\texttt{Name}$, $\tau[\mu][\sigma] \gets \tau[\mu][\sigma] \cup \{\upsilon\}$ 
        \EndIf
      \EndFor
    \EndIf
  \EndFor
\EndFor
\State \Return $\tau$
\end{algorithmic}
\end{algorithm}

\begin{figure}[h]
	\centering
	\includegraphics[width=0.9\linewidth]{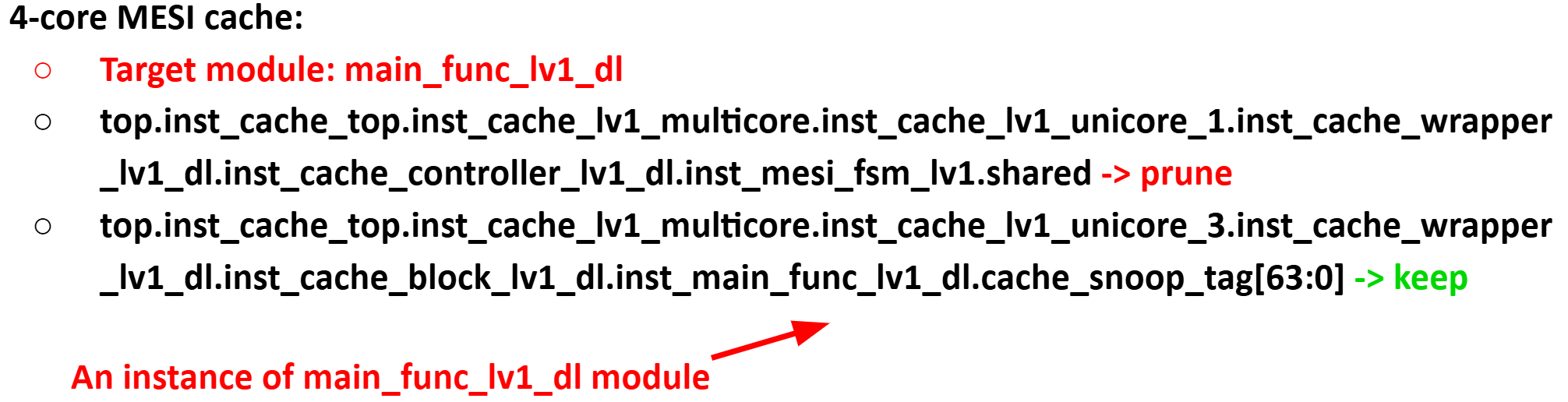}
	\caption{Pruning signals in hierarchical form.}
    \label{fig:crawlerpurpose}    
\end{figure}

 Verible's linter~\cite{verible2024} generates ASTs from Verilog/SystemVerilog code. These ASTs, denoted as $\alpha$, are processed by Algorithm~\ref{algo:crawlandbuild} to construct a module lookup table $\tau$ for a design $\delta$. The algorithm iterates through each module in the AST, extracting the module name $\mu$ and initializing $\tau[\mu]$ to store its variable declarations. It then traverses $\mu$'s subtrees to record \texttt{DataType}s $\sigma$ and corresponding \texttt{RegisterVar}s $\upsilon$. Finally, the signal selector filters the complete hierarchical signal list, retaining signals matching entries in the lookup table $\tau$. Figure~\ref{fig:crawlerpurpose} demonstrates the pruning process.

\subsection{Parallel simulation, extraction, and processing}

 Signal selection by itself is insufficient for designs with numerous modules. Even after pruning, VCD datasets can reach hundreds of gigabytes, making them difficult to process efficiently. To address this challenge, we developed a multi-stage data compression process:

\begin{enumerate}
    \item \textbf{VCDs to CSVs (raw)}: A fixed simulation window of 2000 ticks before termination limits file sizes to around 100 MB. Raw VCDs are then converted to Comma-Separated Values (CSVs), a widely supported tabular format for data processing.

    \item \textbf{Window of failure (rough)}: Some tests terminate earlier than the predefined simulation window, leading to varying CSV lengths. For standardization, we apply a mean window function, trimming longer files and padding shorter ones with zeros.

    \item \textbf{Statistical compression}: We compress each signal into one CSV row using statistical metrics like mean, standard deviation, and quantiles. This shrinks the data size but increases the features. The SummaryTransformer~\cite{sktime_docs} function generates one CSV per bug scenario, with rows as waveforms and $d \times n$ feature columns, where $n$ is the statistical metric count.

    \item \textbf{Final CSV}: All statistically compressed CSVs are packaged into a single final CSV that can be imported into any ML environment.

    \item \textbf{XGBoost feature selection (optional)}: We apply XGBoost for signal ranking based on relevance to training bug scenarios, retaining the top 50\%-70\% of signals and iterating until all modules are included or the signal list reaches a processing limit of 5000 signals.
\end{enumerate}

 Despite extraction-level optimizations, the flow remains time-intensive, particularly for simulation and data processing. Multi-core parallelization is paramount for further performance improvement. 

 We first divided the process into two pipelines: simulation and VCD extraction, and data processing. Each pipeline was further broken down into atomic processes. A dispatcher manages instances of these pipelines, or workers. It assigns each bug scenario to a worker to run through the entire simulation and VCD extraction pipeline until all raw CSVs are collected. The files are then processed through the data pipeline to generate the final CSV file containing compact tabular data.

\subsection{LLM-based synthetic bug generator}

 A major challenge in evaluating our framework is the scarcity of publicly available RTL-level bug datasets. To address this, inspired by recent advances in Large Language Model (LLM) in source code mutation~\cite{10.1109/ICSE48619.2023.00194, ibrahimzada2024automatedbuggenerationera}, we developed an LLM-driven bug generation tool for Verilog/SystemVerilog~\cite{jasper2025buggen}.

\begin{figure}[h]
  \centering
  \includegraphics[width=0.9\linewidth]{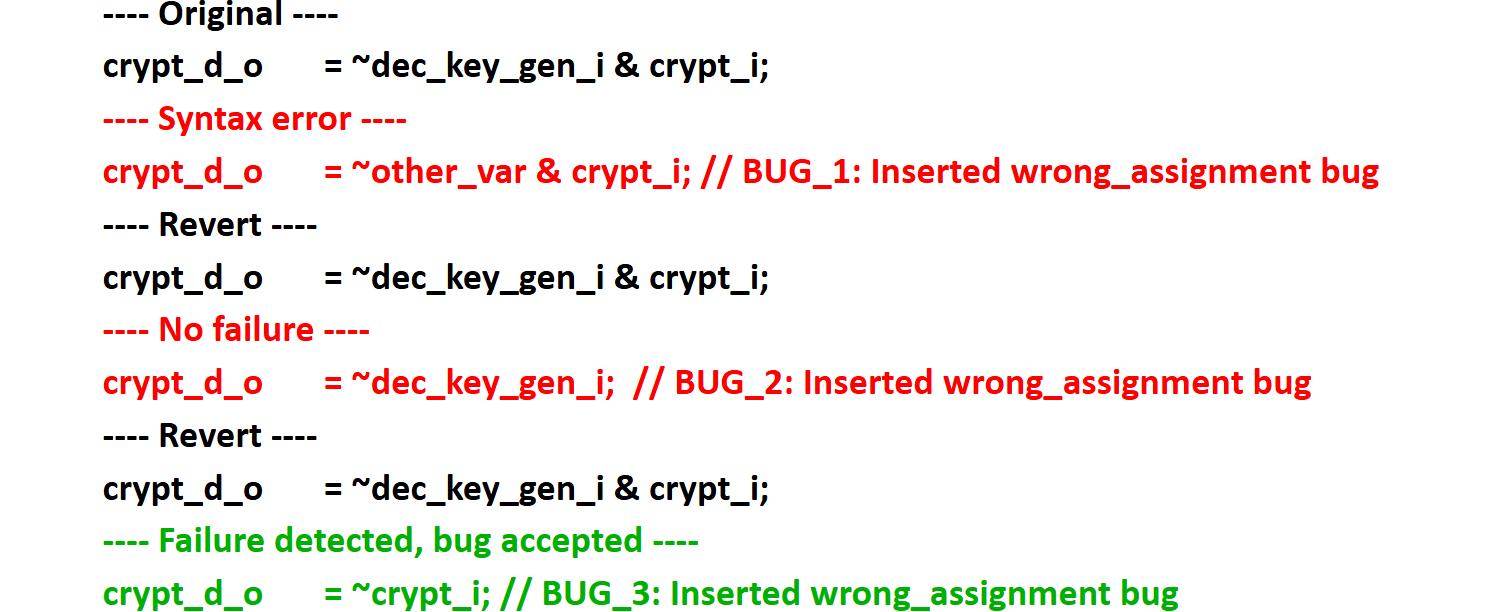}
  \caption{LLM attempts to get an accepted bug scenario.}
  \label{fig:llm_operations}
\end{figure}

 LLMs lack RTL-specific knowledge and struggle with complex designs beyond simple constructs like FSMs or FIFOs. Their instability and limited controllability hinder the autonomous generation of syntactically valid mutations with targeted failure rates. To address this, we decomposed mutation into modular LLM-driven steps: module splitting, region selection, and mutation. A rulebook (Table~\ref{tab:buginsertion}) ensures consistent bug insertion. The system auto-reverts failed attempts, due to syntax errors or ineffective bugs, caches valid mutations, and logs repeated failures to minimize manual effort. Figure~\ref{fig:llm_operations} illustrates the workflow.

\begin{table}[h]
  \footnotesize
  \setlength{\tabcolsep}{10pt}
  \renewcommand{\arraystretch}{1.1}
  \begin{center}
    \begin{tabular}{|p{2.8cm}|p{4.2cm}|}
        \hline
        \textbf{Bug Type} & \textbf{Bug Description} \\
        \hline\hline
        Missing assignment    & An assignment statement is commented out. \\
        Wrong assignment      & A left-hand side variable is incorrectly assigned to a specific value or expression. \\
        Bit-wise corruption   & One or more incorrect logical operators are applied to the right-hand side of an assignment. \\
        Logic bug             & An \texttt{if}, \texttt{while}, or \texttt{always\_ff} statement contains incorrect logic or active edge. \\
        Data size             & A mismatch in vector size between the right-hand and left-hand sides. \\
        \hline
    \end{tabular}
  \end{center}
  \caption{Summary of injected bug types and their descriptions.}
  \label{tab:buginsertion}
\end{table}

 Our generator operates unattended reliably. Users only need to specify the target modules and the number of desired bug scenarios. The system supports applying and rolling back multiple mutations per scenario, enhancing dataset diversity. A notable advantage is its ability to describe common bug types in plain English and adapt across different designs with minimal modification. RTL-level mutations are directly applicable to the source code, allowing historical bugs to remain useful for future training.

 Generation speed depends on testbench quality and signal consistency. Since ML models only train on failing waveforms, weak stimuli and minimal checkers prolong simulations. VCDiag requires consistent signal names and discards valid bugs if a parameter change renames a signal. The lack of multi-core support exacerbates lengthy simulation times due to the sequential nature of the process. For context, in the worst-case scenario, generating 1,600 FabScalar~\cite{fabscalar} scenarios took approximately three days. Although this is a one-time cost and still faster than manual injection, further optimization is planned.

\subsection{Machine learning environment}

This study focuses on classical supervised ML models for their speed and effectiveness. Future work will explore deep learning for MTSC and advanced feature encoding to reduce input dimensionality, both of which are supported by our current infrastructure. All experiments used default model parameters to highlight the impact of our data processing pipeline and ensure reproducibility. Notably, data quality and quantity had a greater influence on performance than hyperparameter tuning, which, from our testing, yielded mixed results. VCDiag currently supports four models:

\begin{enumerate}
    \item \textbf{K-Nearest Neighbors}: KNN is a non-parametric algorithm for classification and regression. Though it involves no training phase, it relies on labeled data to predict the class of a new input based on the majority label among its $k$ closest neighbors. We used scikit-learn’s KNN model~\cite{sklearn_api}.
    \item \textbf{Random Forest}: Random Forest is an ensemble learning algorithm that constructs multiple decision trees during training. Each tree is built on a random subset of data and features, and the final prediction is determined by majority voting. We used scikit-learn's RFC model~\cite{sklearn_api}.
    \item \textbf{Advanced Gradient Boosting Trees}: We evaluated two state-of-the-art tree-based methods—XGBoost (XGB)~\cite{xgboost} and LightGBM (LGBM)~\cite{ke2017lightgbm}. XGBoost builds trees sequentially to correct residual errors using a level-wise strategy, while LGBM adopts a leaf-wise growth approach for faster training and lower memory usage.
\end{enumerate}

\section{Experimental Setup}
 In this section, we clarify the experimental setup, including the benchmark designs, simulation environments, and evaluation metrics employed.
 
 This paper presents two main experiments on the OpenTitan AES block and the FabScalar processor in Table~\ref{tab:design_specs}, along with three smaller ones in Table~\ref{tab:small_design_specs}. Modules were selected across different hierarchy levels to study how signal overlap affects framework performance. For instance, choosing both~\textit{AESCipherControl} and its FSM helps observe shared-signal impact.

 Each module includes 80–120 training and 20 testing bug scenarios, injected and extracted independently for strict separation of training and testing data. Running the bug-injected design through a suite of test cases multiple times, or regression reseeds, produces a proportional number of VCDs. The "Modules (Tr/Te)" row lists these counts (e.g., "CiphControl" has 3,435 training and 194 testing VCDs). For AES, constraint-random stimulus generated diverse VCDs via six training and three testing reseeds. FabScalar lacks such variability due to fixed program binaries, so training scenarios ran twice, and testing ones ran once.

\begin{table}[h]
  \footnotesize
  \setlength{\tabcolsep}{4pt}
  \renewcommand{\arraystretch}{1.1}
  \begin{center}
    \begin{tabular}{|p{2.3cm}|p{2.8cm}|p{2.8cm}|}
        \hline
        \textbf{Info}              & \textbf{AES~\cite{opentitan}} & \textbf{FabScalar~\cite{fabscalar}} \\
        \hline\hline
        Type                & IP Block         & Processor           \\
        Language            & SystemVerilog    & Verilog             \\
        \# of RTL Lines     & 14,739           & 20,290              \\
        \# of Signals       & 24,053           & 17,143              \\
        TB Architecture     & UVM              & ISS Reference       \\
        Simulator           & Synopsys VCS~\cite{synopsys_vcs}     & Cadence Xcelium~\cite{cadence_xcelium}     \\
        \hline\hline
        \# of Target Modules     & 7         & 12                  \\
        \# of Chosen Signals     & 1,612     & 4,860               \\
        \hline\hline
        Modules (Tr/Te)          & CiphCtrl (3,435/194)         & FetchStage1 (1,328/118)  \\
                                 & \_\_\_\_ FSM (3,999/503)     & FetchStage2 (1,333/122)  \\
                                 & AESCtr (2,092/226)           & Decode (1,219/120)       \\
                                 & \_\_\_\_ FSM (2,868/328)     & InstructBuff (1,397/118) \\
                                 & RegClear (4,160/407)         & Rename (1,345/123)       \\
                                 & KeyExpand (3,673/396)        & Dispatch (1,330/123)     \\
                                 & MxSingCol (4,116/286)        & IssueQueue (1,017/126)   \\
                                 &                              & RegRead (1,286/111)      \\
                                 &                              & Execution (565/96)       \\
                                 &                              & LdStrUnit (869/105)      \\
                                 &                              & Retire (1,392/117)       \\
                                 &                              & ArchMapTable (942/70)    \\
        \hline
        \textbf{Total Tr/Te}     & \textbf{24,343/2,340}        & \textbf{15,723/1,349}    \\
        \hline
    \end{tabular}
  \end{center}
  \caption{Experimental setups for OpenTitan AES and FabScalar.}
  \label{tab:design_specs}
\end{table}

\begin{table}[h]
  \footnotesize
  \setlength{\tabcolsep}{3.5pt}
  \renewcommand{\arraystretch}{1.1}
  \begin{center}
    \begin{tabular}{|p{2.3cm}|p{1.7cm}|p{1.7cm}|p{1.7cm}|}
      \hline
      \textbf{Info}              & \textbf{OTPCtrl} & \textbf{RomCtrl} & \textbf{USB} \\
      \hline\hline
      Type                       & Controller           & Controller         & Device           \\
      \# of RTL Lines            & 11,927               & 2,558              & 16,240           \\
      \# of Target Modules       & 6                    & 5                  & 8                \\
      \# of Chosen Signals       & 2,009                & 393                & 711              \\
      \hline
    \end{tabular}
    \caption{Additional experiments on OpenTitan IPs.}
    \label{tab:small_design_specs}
  \end{center}
\end{table}

 We use classification metrics from BugMD~\cite{mammo2016bugmd} and extend them for a more detailed performance analysis. Top-1 accuracy measures the percentage of correct top predictions, while top-3 accuracy includes the correct label in the top three predictions. The F1 score balances precision and recall, making it well-suited for our imbalanced dataset, where failures depend on the modules with injected bugs and the test cases used. True Positive Rate (TPR) and False Positive Rate (FPR) capture sensitivity and false alarm rates, and the Area Under the Curve of the Receiver Operating Characteristic (AUC ROC) summarizes the TPR–FPR trade-off. We macro-average TPR, FPR, and AUC~\cite{scikitlearnMulticlassReceiver}, as defined in Equations~\eqref{eq:auc_roc}, to ensure equal treatment of all classes and encourage robust detection.

 To evaluate parallelization benefits, we ran 40 AES bug scenarios through the full parallel simulation, extraction, and processing pipeline (orange section in Figure~\ref{fig:vcdiag_framework}), measuring the end-to-end completion time across different core configurations.

{\footnotesize
\begin{equation}
\begin{aligned}
\text{TPR}_{\text{macro}} &= \frac{1}{M^{*}} \sum_{i=1}^{M^{*}} \frac{\text{TP}_i}{\text{TP}_i + \text{FN}_i}, \\
\text{FPR}_{\text{macro}} &= \frac{1}{M^{*}} \sum_{i=1}^{M^{*}} \frac{\text{FP}_i}{\text{FP}_i + \text{TN}_i}, \\
\text{AUC}_{\text{ROC}} &\approx \sum_{j=1}^{n-1} \frac{(\text{FPR}_{j+1} - \text{FPR}_j)(\text{TPR}_{j+1} + \text{TPR}_j)}{2}
\end{aligned}
\label{eq:auc_roc}
\end{equation}
\begin{center}
{\footnotesize \(^{*}\) number of modules}
\end{center}
}

\section{Results and Discussion}

 Here, we present a comparative analysis of various ML models, highlighting key performance differences and their implications. Selected case studies illustrate how top-performing models identify root causes in real-world debugging scenarios. We also evaluate resource usage across different core configurations, offering insights into the approach’s efficiency.

\subsection{Classification Results}

\begin{table}[h]
  \footnotesize
  \setlength{\tabcolsep}{5pt}
  \renewcommand{\arraystretch}{1.2}
  \begin{center}
    \begin{tabular}{|p{2.0cm}|c|c|c|c|}
      \hline
      \textbf{Model} & \textbf{Top-1} & \textbf{Top-3} & \textbf{F1} & \textbf{ROC\textsubscript{AUC}} \\
      \hline
      \multicolumn{5}{|c|}{\textbf{AES}} \\
      \hline
      KNN  & 0.577 & 0.817 & 0.553 & 0.840 \\
      RF   & 0.887 & 0.961 & 0.873 & 0.962 \\
      XGB  & 0.851 & 0.962 & 0.841 & 0.969 \\
      LGBM & 0.869 & 0.970 & 0.869 & 0.978 \\
      \hline
      \multicolumn{5}{|c|}{\textbf{FabScalar}} \\
      \hline
      KNN  & 0.593 & 0.766 & 0.586 & 0.839 \\
      RF   & 0.725 & 0.894 & 0.725 & 0.961 \\
      XGB  & 0.760 & 0.936 & 0.757 & 0.974 \\
      LGBM & 0.770 & 0.941 & 0.770 & 0.974 \\
      \hline
      \multicolumn{5}{|c|}{\textbf{OpenTitan IPs with LGBM}} \\
      \hline
      OTPCtrl & 0.898 & 0.958 & 0.885 & 0.986 \\
      ROMCtrl & 0.928 & 0.985 & 0.921 & 0.993 \\
      USB     & 0.932 & 0.979 & 0.939 & 0.998 \\
      \hline
      \textbf{LGBM Avg} & \textbf{0.879} & \textbf{0.967} & \textbf{0.877} & \textbf{0.986} \\
      \hline
    \end{tabular}
  \end{center}
  \caption{Summary of model performances.}
  \label{tab:roc_auc_results}
\end{table}

 Table~\ref{tab:roc_auc_results} summarizes model performances. Tree-based models like XGBoost, LightGBM (LGBM), and Random Forest show varying results due to differences in how they handle high-dimensional, noisy data typical of VCD-derived features. LGBM and XGBoost nearly always outperform Random Forest, thanks to their gradient-based optimization, regularization, and efficient handling of feature interactions. LGBM, in particular, leverages leaf-wise growth and histogram-based binning, making it well-suited for sparse and categorical data. In contrast, Random Forest, though robust to overfitting, lacks such fine-grained optimization and struggles with subtle interactions across many signals. Increasing the number of statistical features further enhances tree-based model accuracy by enabling more informative hierarchical splits. LGBM also benefits from fast training (under five minutes) and rapid inference, making it ideal for our use case. Meanwhile, KNN underperforms due to its reliance on distance metrics, which are less effective in high-dimensional spaces.

\begin{figure}[h]
  \centering
  \includegraphics[width=0.70\columnwidth]{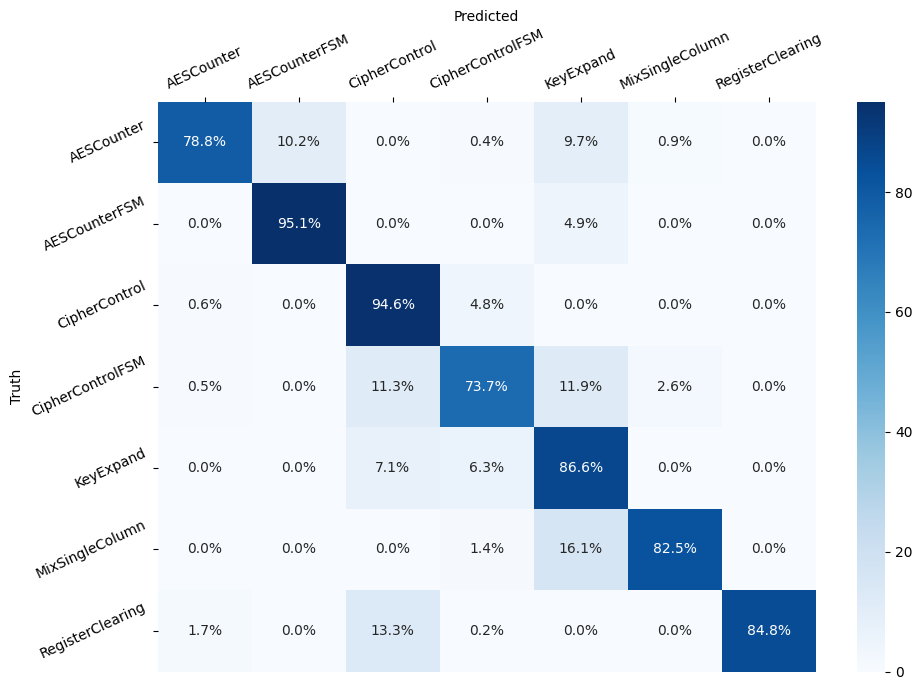}
  \caption{LightGBM Confusion Matrix for AES.}
  \label{fig:lgbmperf_aes_cm}
\end{figure}

\begin{figure}[h]
  \centering
  \includegraphics[width=0.70\columnwidth]{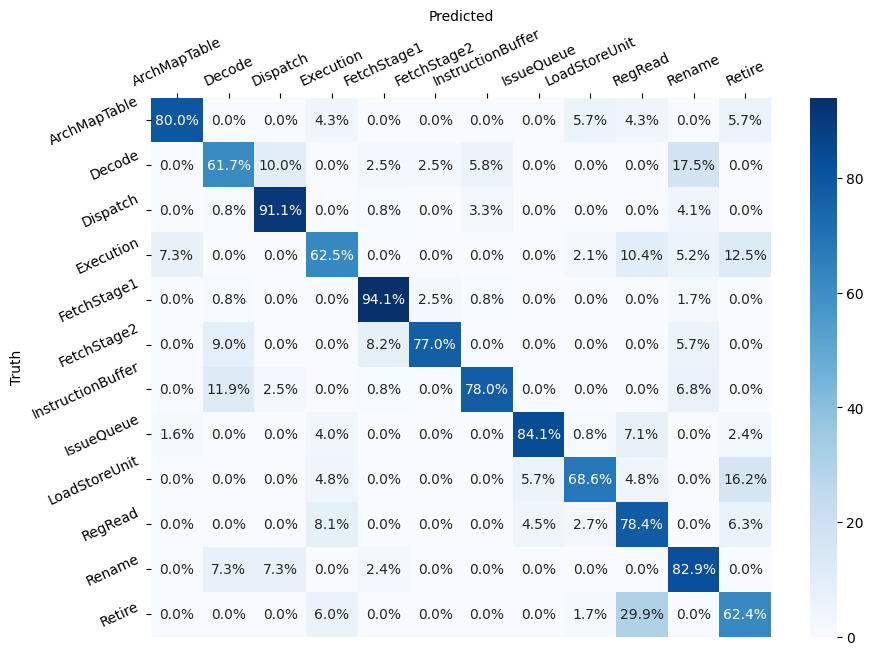}
  \caption{LightGBM Confusion Matrix for FabScalar.}
  \label{fig:lgbmperf_fab_cm}
\end{figure}

 To gain better insight, we examine LGBM confusion matrices in Figure~\ref{fig:lgbmperf_aes_cm} and~\ref{fig:lgbmperf_fab_cm}. For AES, misclassifications frequently occur between \textit{AESCounter} and \textit{AESCounterFSM}, likely due to overlapping signals. Nevertheless, these predictions help narrow the debug scope and accelerate triage. Additionally, the model disproportionately predicts \textit{CipherControl}, \textit{CipherControlFSM}, and \textit{KeyExpand}. According to OpenTitan~\cite{opentitan}, these modules play central roles in AES operation—\textit{CipherControl} governs the cipher core, and \textit{KeyExpand} generates encryption keys. Their frequent activity likely imprints strong statistical signatures, causing misattributions from unrelated modules.

 The performance gap between AES and FabScalar is largely due to testbench complexity. For FabScalar, we initially trained on 70 and tested on 20 bug scenarios, achieving 73\% top-1 accuracy with LGBM. Increasing the training set to 120 scenarios improved accuracy to 77\%, but at a higher generation cost, as the testbench and stimulus limitations hindered the creation of new bug scenarios. The static program trace limits processor behavior diversity, reflecting real-world engineering constraints where limited input variability makes accurate diagnosis more challenging. These findings underscore the importance of testbench design in shaping ML training outcomes and highlight opportunities to refine our training strategies to reduce dependency on handcrafted scenarios.

\subsection{Resource utilization results}

\begin{table}[h]
  \centering
  \footnotesize
  \setlength{\tabcolsep}{2pt} 
  \renewcommand{\arraystretch}{1.1} 
  \begin{tabular}{|p{2.5cm}|p{2.7cm}|p{2.7cm}|}
    \hline
    \textbf{Processing Stage} & \textbf{OpenTitan AES} & \textbf{FabScalar} \\  
    \hline \hline
    Sim and extract   & 14:38:09 & 28:48:18 \\
    Data processing   & 00:52:27 & 02:54:22 \\
    LGBM data loading & 00:01:34 & 00:05:48 \\
    LGBM training     & 00:01:48 & 00:05:55 \\
    LGBM inference    & 00:00:01 & 00:00:01 \\
    \hline \hline
    Raw          & 100 GB  & 308 GB \\
    Rough        & 64  GB  & 248 GB \\
    Compressed   & 4.8 GB  & 8.4 GB \\
    Final CSV    & 2   GB  & 2.5 GB \\
    \hline
    
  \end{tabular}
  \caption{End-to-end timings (in HH:MM:SS) on 32 cores and total data size at each stage of the data processing pipeline.}
  \label{tab:data_compression}
\end{table}

 Efficient storage and compute utilization are pivotal to scalability. Table~\ref{tab:data_compression} shows that waveform compression achieves 50x and 123x data size reductions for AES and FabScalar, lowering storage overhead and speeding up data loading and model inference. Such compression is crucial for handling growing waveform datasets in larger designs, ensuring practical runtimes.

\begin{figure}[h]
  \centering
  \begin{subfigure}[t]{0.65\columnwidth} 
    \centering
    \includegraphics[width=\textwidth]{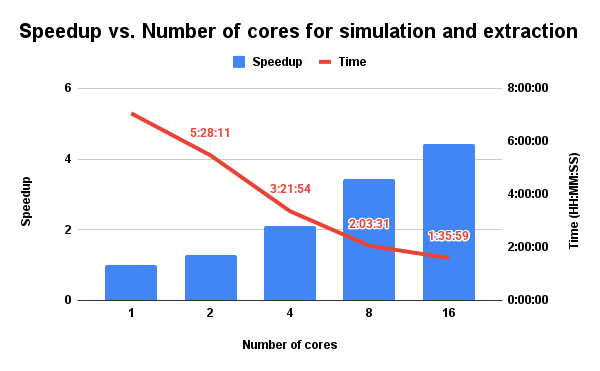}
    \caption{Sim and extract}
    \label{fig:speedup_extract_proc}
  \end{subfigure}
  \hfill
  \begin{subfigure}[t]{0.65\columnwidth} 
    \centering
    \includegraphics[width=\textwidth]{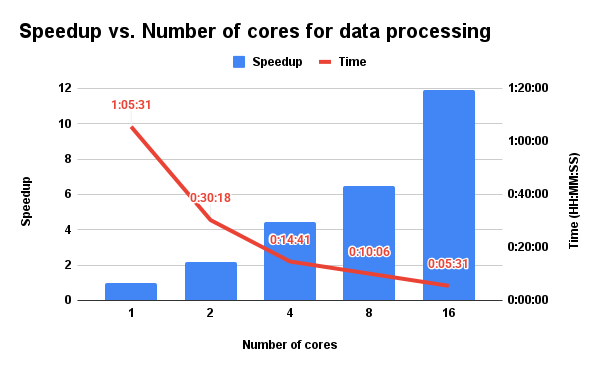}
    \caption{Data processing}
    \label{fig:speedup_data_proc}
  \end{subfigure}
  \caption{Multi-core completion time and speedup for 40 AES' bug scenarios.}
  \label{fig:speedup_proc}
\end{figure}

 Figure~\ref{fig:speedup_extract_proc} shows a 4.4x speedup from 1 to 16 cores. Stalls occur when all cores are occupied, and uneven bug scenario distribution leaves some cores idle near the end of a job. More workers also adds overhead from simulation setup and resource management. The data processing pipeline in Figure~\ref{fig:speedup_data_proc} scales more linearly due to near-complete parallelization, with sequential overheads amortized over larger bug scenario volumes. Final end-to-end timing results are presented for both designs in Table~\ref{tab:data_compression}.

 The AES testbench enables early failure detection, fast simulation, efficient extraction, and compact VCDs. In contrast, FabScalar bug scenarios often lead to stalled commits, with VCDs reaching the 2,000-tick cap but offering limited value for ML training.

\begin{figure}[h]
  \centering
  \includegraphics[width=0.65\columnwidth]{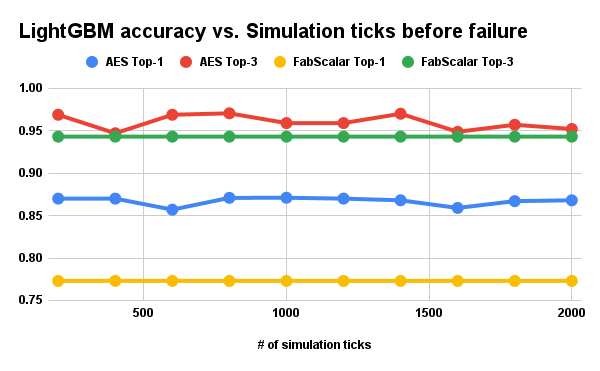}
  \caption{LightGBM accuracy relative to simulation ticks before failure.}
  \label{fig:lgbmperf_vs_cap}
\end{figure}

Despite the 2000-tick cap, Figure~\ref{fig:lgbmperf_vs_cap} shows comparable accuracy at 200 ticks, a major efficiency gain opportunity. This is especially evident in FabScalar, where the marginal difference is likely due to many failures being timeouts, allowing the model to identify the faulty module with a small number of ticks.

\subsection{Adaptation}

 In Figure~\ref{fig:vcdiag_framework}, most framework changes occur between insertion and VCD extraction. This section is self-contained - adding a new design only requires one automation script and a config file. The bug generation flow mirrors insertion and extraction, streamlining script development. Core components like synthetic bug generation, data processing, and the ML environment remain consistent, highlighting the benefits of standardized data and modular framework design.

 Initial OpenTitan RomCtrl integration took three days due to testbench and post-simulation handling, but extending to additional OpenTitan designs required only hours due to shared UVM infrastructure. For FabScalar, the setup required two days to adapt the framework to its distinct flow. By open-sourcing VCDiag, we aim to foster broader adoption and evolution of our framework in both research and industry.

\subsection{Scalability}

 The proposed methodology is inherently scalable to large commercial designs, even though our experiments focus on FabScalar and OpenTitan as representative benchmarks. This scalability is supported by: (a) the use of industrial-grade verification tools and (b) the adherence to industry-standard methodologies like UVM, allowing seamless integration into commercial workflows; (c) reliance on the standard VCD format, which facilitates interoperability with a wide range of industry simulators; (d) dataset reduction techniques specifically designed to handle high-dimensional, large-scale data efficiently; and (e) the demonstrated model flexibility and robustness suggest the approach remains effective for complex, realistic industrial-scale bugs.

\subsection{Future Work}

 VCDiag represents a significant step toward developing a production-ready, ML-based failure triage system, opening several promising avenues for future enhancements. First, leveraging standardized logs with support from LLMs and automated log templates will greatly improve log clustering, thereby enhancing coarse-grain debugging by efficiently grouping related failures. Additionally, extending VCDiag to support scenarios with multiple concurrent faulty modules through an iterative 'onion-peeling' methodology will significantly broaden its practical applicability. This iterative refinement, however, would require developing strategies to accommodate intermediate design modifications seamlessly. Finally, focusing on improving the efficiency of regression reseeds, either through optimized simulation strategies or intelligent prioritization of test scenarios, will accelerate bug dataset generation and bring VCDiag closer to an industry-ready ML debugging framework.

\section{Conclusion} 
 As chips get more complex, resolving debugging bottlenecks, with a focus on failure triage, is critical for improving DFV efficiency and ensuring timely project completion. This research introduces VCDiag, a data mining and ML framework that analyzes failing VCD waveforms to predict suspect modules in Verilog/SystemVerilog designs, providing a reliable and generalizable failure triage method. VCDiag combines an intuitive, structured approach to waveform extraction and processing, emulating standard debugging methodologies, with high prediction accuracy and reasonable end-to-end processing times. The generated datasets are lightweight, ML-ready, and structured clearly for future applications. While challenges remain in handling large VCD files, mitigating simulation bottlenecks, and managing multi-bug triage, the promising results of this research pave the way for further advancements and a more efficient, scalable debugging process.

\section{Acknowledgment}
This work was partially supported by the National Science Foundation (CCF-2106725) and the Hans Fischer Senior Fellowship by the Institute for Advanced Study of the Technical University of Munich.

\pagebreak

\bibliographystyle{IEEEtran}
\bibliography{main}

\begin{thebibliography}{10}
\providecommand{\url}[1]{#1}
\csname url@samestyle\endcsname
\providecommand{\newblock}{\relax}
\providecommand{\bibinfo}[2]{#2}
\providecommand{\BIBentrySTDinterwordspacing}{\spaceskip=0pt\relax}
\providecommand{\BIBentryALTinterwordstretchfactor}{4}
\providecommand{\BIBentryALTinterwordspacing}{\spaceskip=\fontdimen2\font plus
\BIBentryALTinterwordstretchfactor\fontdimen3\font minus \fontdimen4\font\relax}
\providecommand{\BIBforeignlanguage}[2]{{%
\expandafter\ifx\csname l@#1\endcsname\relax
\typeout{** WARNING: IEEEtran.bst: No hyphenation pattern has been}%
\typeout{** loaded for the language `#1'. Using the pattern for}%
\typeout{** the default language instead.}%
\else
\language=\csname l@#1\endcsname
\fi
#2}}
\providecommand{\BIBdecl}{\relax}
\BIBdecl

\bibitem{verifgap2024}
H.~D. Foster, ``The 2024 wilson research group functional verification study,'' 2024.

\bibitem{1512377}
A.~Smith, A.~Veneris, M.~Ali, and A.~Viglas, ``Fault diagnosis and logic debugging using boolean satisfiability,'' \emph{IEEE Transactions on Computer-Aided Design of Integrated Circuits and Systems}, vol.~24, no.~10, pp. 1606--1621, 2005.

\bibitem{6604054}
Z.~Poulos, Y.-S. Yang, and A.~Veneris, ``A failure triage engine based on error trace signature extraction,'' in \emph{2013 IEEE 19th International On-Line Testing Symposium (IOLTS)}, 2013, pp. 73--78.

\bibitem{fae}
C.-H. Shen, A.~C.-W. Liang, C.~C.-H. Hsu, and C.~H.-P. Wen, ``Fae: Autoencoder-based failure binning of rtl designs for verification and debugging,'' in \emph{2019 IEEE International Test Conference (ITC)}, 2019, pp. 1--10.

\bibitem{truong2018clustering}
A.~Truong, D.~Hellstr{\"o}m, H.~Duque, and L.~Viklund, ``Clustering and classification of uvm test failures using machine learning techniques,'' in \emph{Proceedings of the Design and Verification Conference (DVCON), San Jose, CA, USA}, vol.~26, 2018.

\bibitem{safarpour2012failure}
S.~Safarpour, B.~Keng, Y.-S. Yang, and E.~Qin, ``Failure triage: The neglected debugging problem,'' in \emph{Design and Verification Conference}, 2012.

\bibitem{mammo2016bugmd}
B.~Mammo, M.~Furia, V.~Bertacco, S.~Mahlke, and D.~S. Khudia, ``Bugmd: Automatic mismatch diagnosis for bug triaging,'' in \emph{2016 IEEE/ACM International Conference on Computer-Aided Design (ICCAD)}.\hskip 1em plus 0.5em minus 0.4em\relax IEEE, 2016, pp. 1--7.

\bibitem{verilogstandard}
``Ieee standard hardware description language based on the verilog(r) hardware description language,'' \emph{IEEE Std 1364-1995}, pp. 1--688, 1996.

\bibitem{verible2024}
``Verible,'' \url{https://github.com/chipsalliance/verible}, 2024.

\bibitem{vcdvcd2024}
cirosantilli, ``Vcdvcd,'' \url{https://github.com/cirosantilli/vcdvcd}, 2024.

\bibitem{pandas}
W.~McKinney, ``Data structures for statistical computing in python,'' in \emph{Proceedings of the 9th Python in Science Conference}, S.~van~der Walt and J.~Millman, Eds., 2010, pp. 51 -- 56.

\bibitem{opentitan}
``Opentitan: Open source silicon root of trust (rot),'' \url{https://github.com/lowrisc/opentitan}, 2024.

\bibitem{fabscalar}
N.~K. Choudhary, S.~V. Wadhavkar, T.~A. Shah, H.~Mayukh, J.~Gandhi, B.~H. Dwiel, S.~Navada, H.~H. Najaf-abadi, and E.~Rotenberg, ``Fabscalar: Composing synthesizable rtl designs of arbitrary cores within a canonical superscalar template,'' in \emph{2011 38th Annual International Symposium on Computer Architecture (ISCA)}, 2011, pp. 11--22.

\bibitem{xgboost}
\BIBentryALTinterwordspacing
T.~Chen and C.~Guestrin, ``Xgboost: A scalable tree boosting system,'' in \emph{Proceedings of the 22nd ACM SIGKDD International Conference on Knowledge Discovery and Data Mining}, ser. KDD '16.\hskip 1em plus 0.5em minus 0.4em\relax New York, NY, USA: Association for Computing Machinery, 2016, p. 785–794. [Online]. Available: \url{https://doi.org/10.1145/2939672.2939785}
\BIBentrySTDinterwordspacing

\bibitem{sktime_docs}
sktime developers, ``sktime documentation,'' \url{https://www.sktime.net/en/stable/}, 2025, accessed: 2025-07-10.

\bibitem{10.1109/ICSE48619.2023.00194}
\BIBentryALTinterwordspacing
S.~Kang, J.~Yoon, and S.~Yoo, ``Large language models are few-shot testers: Exploring llm-based general bug reproduction,'' in \emph{Proceedings of the 45th International Conference on Software Engineering}, ser. ICSE '23.\hskip 1em plus 0.5em minus 0.4em\relax IEEE Press, 2023, p. 2312–2323. [Online]. Available: \url{https://doi.org/10.1109/ICSE48619.2023.00194}
\BIBentrySTDinterwordspacing

\bibitem{ibrahimzada2024automatedbuggenerationera}
\BIBentryALTinterwordspacing
A.~R. Ibrahimzada, Y.~Chen, R.~Rong, and R.~Jabbarvand, ``Automated bug generation in the era of large language models,'' 2024. [Online]. Available: \url{https://arxiv.org/abs/2310.02407}
\BIBentrySTDinterwordspacing

\bibitem{jasper2025buggen}
\BIBentryALTinterwordspacing
S.~Jasper, M.~Luu, E.~Pan, A.~Tyagi, M.~Quinn, J.~Hu, and D.~K. Houngninou, ``Buggen: A self-correcting multi-agent llm pipeline for realistic rtl bug synthesis,'' 2025. [Online]. Available: \url{https://arxiv.org/abs/2506.10501}
\BIBentrySTDinterwordspacing

\bibitem{sklearn_api}
L.~Buitinck, G.~Louppe, M.~Blondel, F.~Pedregosa, A.~Mueller, O.~Grisel, V.~Niculae, P.~Prettenhofer, A.~Gramfort, J.~Grobler, R.~Layton, J.~VanderPlas, A.~Joly, B.~Holt, and G.~Varoquaux, ``{API} design for machine learning software: experiences from the scikit-learn project,'' in \emph{ECML PKDD Workshop: Languages for Data Mining and Machine Learning}, 2013, pp. 108--122.

\bibitem{ke2017lightgbm}
G.~Ke, Q.~Meng, T.~Finley, T.~Wang, W.~Chen, W.~Ma, Q.~Ye, and T.-Y. Liu, ``Lightgbm: A highly efficient gradient boosting decision tree,'' \emph{Advances in neural information processing systems}, vol.~30, 2017.

\bibitem{synopsys_vcs}
{Synopsys, Inc.}, ``{VCS: Functional Verification Solution},'' 2025.

\bibitem{cadence_xcelium}
{Cadence Design Systems, Inc.}, ``{Xcelium Logic Simulation},'' 2025.

\bibitem{scikitlearnMulticlassReceiver}
\BIBentryALTinterwordspacing
``Multiclass receiver operating characteristic (roc),'' 2025, accessed: 2025-05-19. [Online]. Available: \url{https://scikit-learn.org}
\BIBentrySTDinterwordspacing

\end{thebibliography}

\newpage
\section*{Artifact Appendix}

\subsection{Abstract}

This artifact supports two papers: (1) \textbf{VCDiag: Classifying Erroneous Waveforms for Failure Triage Acceleration}, and (2) \textbf{BugGen: A Self-Correcting Multi-Agent LLM Pipeline for Realistic RTL Bug Synthesis~\cite{jasper2025buggen}}.
 It includes code, datasets, and workflows for bug injection, waveform extraction, and ML-based classification.

\subsection{Artifact check-list (meta-information)}

{\small
\begin{itemize}
  \item {\bf Algorithms: } LLM-guided RTL bug injection, ML-based waveform classification
  \item {\bf Program: } Python scripts for bug injection, simulation, extraction, and ML training/inference
  \item {\bf Model: } GPT-4o mini for RTL bug injection, ML models listed for waveform classification
  \item {\bf Dataset: } VCD Raw ASCII waveform datasets from OpenTitan and FabScalar
  \item {\bf Run-time environment: } Linux, Python 3.12
  \item {\bf Hardware: } Multi-core CPU and GPU recommended
  \item {\bf Execution: } Shell scripts and Python workflows
  \item {\bf Metrics: } Classification accuracy, bug injection success rate
  \item {\bf Output: } Classified waveform results, logs, trained models
  \item {\bf Experiments: } Bug injection, waveform extraction, ML training/inference
  \item {\bf Publicly available: } Yes
  \item {\bf Code licenses: } MIT License
  \item {\bf Data licenses: } CC BY 4.0
  \item {\bf Archived: }
    \begin{itemize}
      \item {\bf Designs:} \url{https://zenodo.org/records/16735572}
      \item {\bf Waveform datasets:} \url{https://zenodo.org/records/16791965}
    \end{itemize}
\end{itemize}
}

\subsection{Description}

The artifact consists of two main stages:

\begin{itemize}
  \item \textbf{Bug Injection and VCD Extraction:}  
  Uses LLMs to inject bugs into RTL designs and extract waveform data. The flow is configuration-driven and supports multiple modes for bug generation and waveform extraction. It is scalable and adaptable to different designs.

  \item \textbf{VCD Processing and ML Training/Inference:}  
  Converts raw waveform data into tabular format through cleaning and statistical compression. The processed data is used to train ML models for classification. The flow is parallelized and design-agnostic.
\end{itemize}

The data structure includes folders for raw, cleaned, compressed, and final train/test datasets. The ML pipeline is implemented in Jupyter Notebooks and supports reproducible experiments.

\subsection{Installation and Usage}

For detailed setup instructions, environment configuration, and command-line usage, please refer to the GitHub repository:

\url{https://github.com/minhluu2000/mlcad2025\_vcdiag}

\subsection{Methodology}

Submission, reviewing, and badging methodology:

\begin{itemize}
  \item \url{https://www.acm.org/publications/policies/artifact-review-and-badging-current}
  \item \url{http://cTuning.org/ae/submission-20201122.html}
  \item \url{https://github.com/ml-eda/artifact-evaluation/}
\end{itemize}

\end{document}